\documentclass[a4paper,11pt,twocolumn,twoside]{article}
\usepackage{sepln_en}
\usepackage{fullname}
\usepackage[utf8]{inputenc}
\usepackage[english]{babel}

\usepackage{graphicx}
\usepackage{multirow}
\usepackage{booktabs}
\usepackage{amssymb}
\usepackage{float}
\usepackage{xcolor, colortbl}
\usepackage{adjustbox}
\usepackage{amsmath}
\usepackage{url}

\usepackage[T1]{fontenc}

\input epsf
\title{Multilingual Controllable Transformer-Based Lexical Simplification}
\seplntranstitle{Simplificación Léxica Controlable Multilingüe con Transformers}

\author {\textbf{Kim Cheng Sheang}, \textbf{Horacio Saggion}\\
LaSTUS Group, TALN Lab, DTIC\\
Universitat Pompeu Fabra\\
\{kimcheng.sheang, horacio.saggion\}@upf.edu\\
}

\seplnabstract{Text is by far the most ubiquitous source of knowledge and information and should be made easily accessible to as many people as possible; however, texts often contain complex words that hinder reading comprehension and accessibility. Therefore, suggesting simpler alternatives for complex words without compromising meaning would help convey the information to a broader audience. This paper proposes mTLS, a multilingual controllable Transformer-based Lexical Simplification (LS) system fined-tuned with the T5 model. The novelty of this work lies in the use of language-specific prefixes, control tokens, and candidates extracted from pre-trained masked language models to learn simpler alternatives for complex words. The evaluation results on three well-known LS datasets -- LexMTurk, BenchLS, and NNSEval -- show that our model outperforms the previous state-of-the-art models like LSBert and ConLS. Moreover, further evaluation of our approach on the part of the recent TSAR-2022 multilingual LS shared-task dataset shows that our model performs competitively when compared with the participating systems for English LS and even outperforms the GPT-3 model on several metrics. Moreover, our model obtains performance gains also for  Spanish and Portuguese.}
\seplnkey{Multilingual Lexical Simplification, Controllable Lexical Simplification, Text Simplification, Multilinguality.}
\seplnresumen{Los textos son la fuente más extendida de transferencia de conocimiento e información y deberían ser accesibles a todos. Sin embargo, los textos pueden contener palabras difíciles de entender, viéndose limitada su accesibilidad. En consecuencia, la substitución de palabras difíciles por alternativas más simples, que por otro lado no comprometan el sentido original del texto, podría ayudar a hacer la información más fácil de entender. En este trabajo proponemos el sistema mTLS de simplificación léxica multilingüe controlable basado en “transformers” multilingües, del tipo T5.   La novedad de nuestro método consiste en combinar prefijos específicos del idioma, tokens de control y candidatos extraídos de modelos de lenguaje enmascarados pre-entrenados. Los resultados  obtenidos por mTLS en tres conjuntos de datos para el inglés, muy conocidos en simplificación léxica -- LexMTurk, BenchLS, and NNSEval --  indican que mTLS se comporta mejor que el estado del arte. Además, una evaluación adicional sobre una parte de los datos de la reciente evaluación TSAR-2022 (para simplificación léxica en inglés, español, y portugués) muestra que nuestro modelo supera a todos los sistemas que participaron en la tarea TSAR-2022 en inglés, incluido un modelo basado en GPT-3.  Nuestros resultados para español y portugués indican que mTLS funciona mejor que todos los resultados enviados a  TSAR-2022. }
\seplnclave{Simplificación léxica multilingüe, simplificación de Texto, Simplificación léxica controlable, multilingüismo.}

\firstpageno{1}

\begin{document}


\setlength\titlebox{22.5cm} 

\label{firstpage} \maketitle



\section{Introduction}
    
    Lexical Simplification (LS) is a process of reducing the lexical complexity of a text by replacing difficult words with simpler substitutes or expressions while preserving its original information and meaning \cite{shardlow_survey_2014}. For example, in Figure \ref{figure:an_ls_example}, the word ``motive'' is selected as a complex word, which is replaced by the word ``reason''. Meanwhile, simplification can also be carried out at the syntax level, reducing a text's syntactic complexity. The task is called Syntactic Simplification (SS). Both LS and SS tasks are commonly used as sub-tasks of the broader task of Automatic Text Simplification \cite{saggion_automatic_2017}, which reduces the text's lexical and syntactic complexity. LS systems are commonly composed of different combinations of components such as 1) complex word identification; 2) substitute generation or extraction; 3) substitute filtering; 4) substitute ranking; and 5) morphological and contextual adaptation \cite{paetzold_survey_2017}. 

    \begin{figure}[t]
        \centering
        \def\arraystretch{1.3}
        \begin{tabular}{|p{0.95 \linewidth}|}
        \hline
        The \textbf{motive} for the killings was not known.
         \\ 
        $ \ \ \ \ \ \ \ \ \ \ \Downarrow$ \\     
        The \textbf{reason} for the killings was not known.\\
        \hline
        \end{tabular}
        \caption{A lexical simplification example taken from the TSAR English dataset \cite{saggion-etal-2022-findings} with the complex word and the substitute word in bold.}
        \label{figure:an_ls_example}
    \end{figure}
    Previous works on LS have relied on an unsupervised approach \cite{Biran:11,horn_learning_2014,glavas_simplifying_2015}, and many other systems are module based \cite{ferres_adaptable_2017,gooding_recursive_2019,alarcon_exploration_2021}, which requires a pipeline of modules to operate, such as substitute generation, substitution selection, substitution filtering, and substitution ranking. The downside of the pipeline approach is that it is known to propagate errors across modules. 


    In \namecite{sheang-etal-2022-controllable}, we proposed an end-to-end controllable LS system. However, this model lacks multilinguality; therefore, here we extend that work to show how it can be ported to other languages by jointly learning different languages simultaneously.  
    
    We present the following contributions:
    
    \begin{itemize}
        \item We improve the English monolingual LS model and propose a new multilingual LS model for English, Spanish, and Portuguese\footnote{The source code and data are available at \url{https://www.github.com/kimchengsheang/mTLS}}. 
        \item We show the way to fine-tune a multilingual LS model by adding language-specific prefixes, control tokens, and Masked Language Model (MLM) candidates extracted from BERT-based pre-trained models. 
        \item We have conducted an extensive analysis comparing our models with several evaluation metrics, which allows us to capture the strengths and weaknesses of our approach.
        
    \end{itemize}
    
    The rest of the paper is organized as follows: Section \ref{sec: related work} presents some related work on Lexical Simplification. 
    Section \ref{sec: system description} explains our proposed model in detail. 
    Section \ref{sec: experiments} describes all the datasets being used, the baselines, the evaluation metrics, how the data is prepared, and the experimental setup.
    Section 5 discusses the results of the experiments, while Section 6 concludes the paper.

\section{Related Work} \label{sec: related work}

    Prior works on Lexical Simplification were mainly based on unsupervised approaches. \namecite{Belder2010} used Latent Words Language Models to reduce text complexity for children. \namecite{horn_learning_2014} proposed a Support Vector Machines (SVM) model trained on an automatically aligned between normal Wikipedia and simple Wikipedia text. \namecite{glavas_simplifying_2015} proposed an approach that utilized GloVe embeddings \cite{pennington_glove_2014} for candidate generation and ranked different features extracted from language models and word frequency. 

    \namecite{qiang_lsbert_2021} proposed LSBert, a LS system that uses Masked Language Model (MLM) approach to extract candidates from BERT pre-trained model \cite{devlin_bert_2019} and rank them by different features such as MLM probability, word frequency, language model, similarity (FastText cosine similarity), and PPDB data \cite{Vulic2018}. 
    
    \namecite{martin_controllable_2019} was the first to introduce ACCESS, a Controllable Sentence Simplification system based on a sequence-to-sequence model, trained with four tokens: number of characters token, Levenshtein similarity token, Word Rank token (the inverse frequency order from extracted from FastText), and dependency tree depth. These four tokens are used to control different aspects of the output sentences: 1) sentence compression, 2) the amount of paraphrasing, 3) lexical complexity, and 4) syntactical complexity. The approach was later adopted by \namecite{sheang_controllable_2021} fine-tuned with T5 \cite{raffel_T5_exploring_2020}, \namecite{martin_muss_2022} fine-tuned with BART \cite{lewis_bart_2020}, and \namecite{maddela_controllable_2021} fine-tuned larger T5.
    
    In \namecite{sheang-etal-2022-controllable}, we introduced ConLS, the first controllable Lexical Simplification system fine-tuned with T5 using three tokens: Word Length token, Word Rank token, and Candidate Ranking token. The three tokens were used to control different aspects of the generated candidates: Word Length is often correlated with word complexity, Word Rank is the frequency order (word complexity is also correlated with frequency), and Candidate Ranking is for the model to learn how to rank the generated candidates through training. The model was fine-tuned with T5-large on TSAR-EN dataset \cite{saggion-etal-2022-findings} and tested on LexMTurk \cite{horn_learning_2014}, BenchLS \cite{paetzold_benchls_2016}, and NNSeval \cite{paetzold_nnseval_2016}. 
    
    There have been some works on Lexical Simplification for Spanish, namely, \namecite{moreno_lexical_2019} proposed readability and understandability guidelines, \namecite{alarcon_easier_2021} released the EASIER dataset, and \namecite{alarcon_exploration_2021} explored the use of different word embeddings models from complex word identification, to substitute generation, selection, and ranking. 

    In this work, we extend our previous work of ConLS, addressing multilinguality along with adding two new control tokens (Word Syllable and Sentence Similarity) and Masked Language Model candidates to improve the model's performance.

\section{Method} \label{sec: system description}
    
    \begin{figure*}[ht]
        \begin{adjustbox}{width=1\textwidth}
        \centering
        \includegraphics{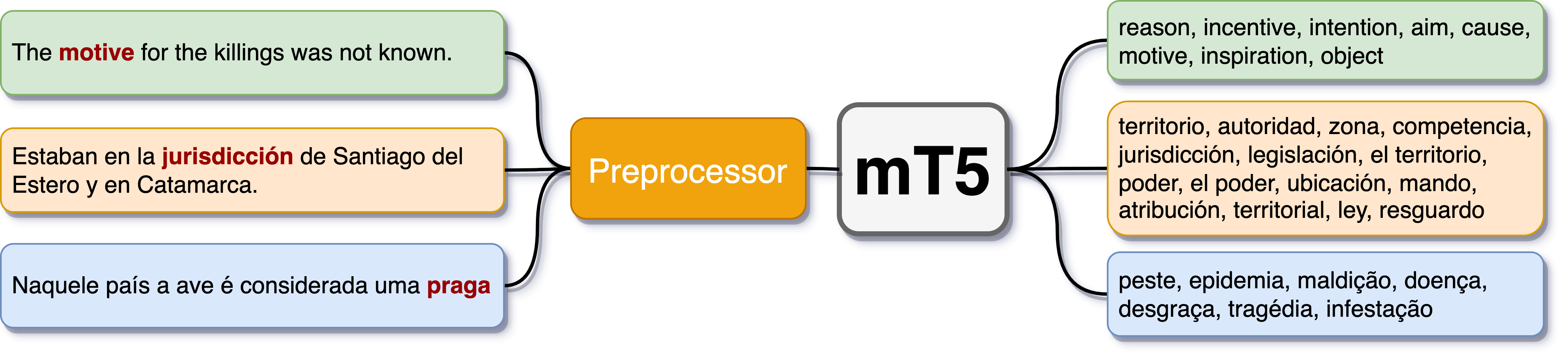}
        \end{adjustbox}
        \caption{Illustration of the mTLS model with three simplification examples from the three languages.}
        \label{figure: model architecture}
        
    \end{figure*}
    
    Building upon the work of ConLS, we propose a new multilingual controllable Transformer-based Lexical Simplification model that integrates language-specific prefixes alongside the control tokens and masked language model candidates to leverage the input-level information. We adopted the same three tokens from ConLS (Word Length, Word Rank, and Candidate Ranking) and integrated two additional tokens (Word Syllables and Sentence Similarity). We fine-tuned our English monolingual model with T5 \cite{raffel_T5_exploring_2020} and multilingual model with mT5 \cite{xue_mt5_2021}. Figure \ref{figure: model architecture} shows an overview of our multilingual model where each input is a sentence with a complex word annotated, and the output is a list of substitutes ranked from the most relevant and simplest to the least. The details of the Preprocessor are described in Section \ref{sec: preprocessing}.  

    \paragraph{Language-specific Prefixes} are embedded into each input so that the model knows and learns to differentiate the three languages. We used three prefixes: ``simplify en:'' for English, ``simplify es:''  for Spanish, and ``simplify pt:'' for Portuguese. In addition, these prefixes serve another purpose. Due to the limited data for Spanish and Portuguese, training individual models for Spanish and Portuguese would make the model unable to generalize well, so to tackle this issue, we jointly trained the three languages in just one model. This way, all the weights are learned and shared between the three languages during the training. 
        
    \paragraph{Control Tokens} The following are the control tokens that were employed in our model to control different aspects of the generated candidates. Word Length, Word Rank (word frequency), and Word Syllables are known to be correlated well with word complexity, so we use them to help select simpler candidates. Candidate Ranking is used to help the model learn how to rank candidates through the training process so that, at the inference, the model could generate and sort candidates automatically, whereas Sentence Similarity is intended to help select relevant candidates based on semantic similarity. 
    
        \begin{itemize}
            \item{\textbf{Word Length (WL)}} is the proportion of character length between a complex word and its substitute. It is calculated by dividing the number of characters in the substitute by the number of characters in the complex word. 

            \item{\textbf{Word Rank (WR)}} is the inverse frequency of the substitute divided by that of the complex word. The frequency order is extracted from the FastText pre-trained model for its corresponding language. Words in FastText pre-trained model are sorted by frequency in descending order\footnote{\url{https://fasttext.cc/docs/en/crawl-vectors.html}}. 

            \item{\textbf{Word Syllables (WS)}} is the ratio of the number of syllables of the substitute divided by that of the complex word. It is extracted using PyHyphen library\footnote{https://github.com/dr-leo/PyHyphen}. The study of \namecite{shardlow_complex_2020} shows that syllable count could help predict lexical complexity.

            \item{\textbf{Candidate Ranking (CR)}} is the ranking order extracted from gold candidates in the training set and normalized to the following values: 1.00 for the first rank, 0.75 for the second rank, 0.5 for the third rank, 0.25 for the fourth rank, and 0.10 for the rest. For the validation set and test set, we set the value to 1.00 for each instance, as we already knew that the best ranking value is 1.00. 

            \item{\textbf{Sentence Similarity (SS)}} is the normalized sentence similarity score between the source and the target sentence. The target sentence is the source sentence with the complex word replaced by its substitute. The score is calculated with the cosine similarity between the embeddings of the two sentences extracted from Sentence-BERT \cite{reimers_sentence-bert_2019,reimers-2020-multilingual-sentence-bert}. This similarity score gives us a measure of the relation between the two sentences. In the experiments, we used the pre-trained model called ``multi-qa-mpnet-base-dot-v1''\footnote{\url{https://www.sbert.net/docs/pretrained_models.html}} because it achieved the best performance on semantic search (tested on 6 datasets) and supported different languages such as English, Spanish, Portuguese, and more. 
            
        \end{itemize}
        
    \paragraph{Masked Language Model (MLM) Candidates}
        The candidates are extracted using the masked language model approach following the same style as LSBert candidates generation. For each input sentence and its complex word, we give the model (e.g., BERT, RoBERTa) the sentence and the same sentence with the complex word masked. E.g., 
        \begin{quote}
            The \textbf{motive} for the killings was not known. </s> The \textbf{[MASK]} for the killings was not known.
        \end{quote}
        We then ask the model to predict the [MASK] token candidates and rank them by the returned probability scores. After that, we select only the top-10 ranked candidates and append them to the end of each input. We believe that adding the MLM candidates to the input sentence could help the model find and select better candidates. More details about how we chose the best pre-trained model for each dataset are described in Section \ref{sec: preprocessing}.

\section{Experiments} \label{sec: experiments}

    In this section, we describe in detail all the datasets, baselines, evaluation metrics, data preparation steps, model details, training, and evaluation procedures. 
    
    \subsection{Datasets}
        
        \begin{table*}[ht]
            \def\arraystretch{1.3}
            \centering
            \begin{tabular}{p{0.7cm} p{4cm} l p{7.5cm}}
            \toprule
                Lang & Text & Target & Ranked Substitutes\\ 
                \midrule
                EN & The motive for the killings was not known.	& motive & reason:16, incentive:2, intention:2, aim:1, cause:1, motive:1, inspiration:1, object:1 \\
                \midrule
                ES & Estaban en la jurisdicción de Santiago del Estero y en Catamarca.	 &	jurisdicción &  territorio:5, autoridad:5, zona:3, competencia:2, jurisdicción:1, legislación:1, el territorio:1, poder:1, el poder:1, ubicación:1, mando:1, atribución:1, territorial:1, ley:1, resguardo:1	\\
                \midrule
                PT & Naquele país a ave é considerada uma praga	 & praga & 	peste:9, epidemia:5, maldição:3, doença:2, desgraça:2, tragédia:1, infestação:1\\
                \hline
            \end{tabular}
            \caption{Three examples from the TSAR-2022 shared-task dataset. Target is the complex word that is already annotated in the datasets. The number after the ``:'' indicates the number of repetitions suggested by crowd-sourced annotators.}
            \label{table:example_from_tsar}
        \end{table*}
        
        In our experiments, we used monolingual English datasets such as LexMTurk \cite{horn_learning_2014}, BenchLS\footnote{\url{https://doi.org/10.5281/zenodo.2552393}} \cite{paetzold_benchls_2016},  NNSeval\footnote{\url{https://doi.org/10.5281/zenodo.2552381}} \cite{paetzold_nnseval_2016}, and a multilingual dataset, TSAR-2022 shared dataset \cite{saggion-etal-2022-findings}. TSAR-2022 dataset contains three subsets: TSAR-EN for English, TSAR-ES for Spanish, and TSAR-PT for Brazilian Portuguese. Table \ref{table:example_from_tsar} shows three examples from the TSAR-2022 dataset, one from each language, and Table \ref{table:datasets-statistics} shows some statistics of the datasets. The average number of tokens (Avg \#Tokens) shows that, on average, TSAR-ES has the longest text length, and TSAR-PT has the shortest text length.  
        
        All datasets that are used in the experiments already have complex words annotated, so the complex word identification module is not needed.

        \begin{table}[h]
        \def\arraystretch{1.2}
        \begin{adjustbox}{width=\columnwidth}
            \begin{tabular}{lccccc}
                \toprule
                \multirow{2}{*}{Dataset} & \multirow{2}{*}{Lang} & \multirow{2}{*}{\#Instances} & \multicolumn{3}{c}{\#Tokens} \\
                \multicolumn{1}{c}{} &  &  & Min & Max & Avg \\ 
                \midrule
                \multirow{3}{*}{TSAR} & EN & 386 & 6 & 83 & 29.85 \\
                 & ES & 381 & 5 & 138 & 35.14 \\
                 & PT & 386 & 3 & 57 & 23.12 \\ \midrule
                LexMTurk & EN & 500 & 6 & 78 & 26.23 \\ \midrule
                BenchLS & EN & 929 & 6 & 100 & 27.90 \\ \midrule
                NNSEval & EN & 239 & 7 & 78 & 27.95 \\ 
                \bottomrule
                \end{tabular}
            \end{adjustbox}
            \caption{Some statistics of the datasets.}
            \label{table:datasets-statistics}
        \end{table}

    \subsection{Baselines}
        
        
        We compare the proposed models with the following strong baselines:
        
        \textbf{LSBert} uses Bert Masked Language Model (MLM) for candidate generation and ranks them by MLM probability, word frequency, language model, similarity (FastText cosine similarity), and PPDB database.
       
        \textbf{ConLS} is a controllable LS system fine-tuned on the T5 model with three control tokens. The candidate generation and ranking are learned through the fine-tuning process. 

        Systems from the TSAR-2022 shared task:
        \begin{itemize}
            \item \textbf{CILS} \cite{seneviratne-etal-2022-cils} generates candidates using language model probability and similarity score and ranks them by candidate generation score and cosine similarity. 
            \item \textbf{PresiUniv} \cite{whistely-etal-2022-presiuniv} uses the Masked Language Model (MLM) for candidate generation and ranks them by cosine similarity and filters using the part-of-speech check. 
            \item \textbf{UoM\&MMU} \cite{vasquez-rodriguez-etal-2022-uom} uses a Language Model with prompts for candidate generation and ranks them by fine-tuning the Bert-based model as a classifier. 
            \item \textbf{PolyU-CBS} \cite{chersoni-hsu-2022-polyu} generates candidates using MLM and ranks them by MLM probability, GPT-2 probability, sentence probability, and cosine similarity. 
            \item \textbf{CENTAL} \cite{wilkens-etal-2022-cental} generate candidates using MLM and ranks them by word frequency and a binary classifier. 
            \item \textbf{teamPN} \cite{nikita-rajpoot-2022-teampn} generates candidates using MLM, VerbNet, PPDB database, and Knowledge Graph and ranks them by MLM probability.
            \item \textbf{MANTIS} \cite{li-etal-2022-mantis} generates candidates using MLM and ranks them by MLM probability, word frequency, and cosine similarity.
            \item \textbf{UniHD} \cite{aumiller-gertz-2022-unihd} uses prompts with GPT-3 (few-shot learning) for candidate generation and ranks them by aggregating the results. 
            \item \textbf{RCML} \cite{aleksandrova-brochu-dufour-2022-rcml} generates candidates using lexical substitution and ranks them by part of speech, BERTScore, and SVM classifier. 
            \item \textbf{GMU-WLV} \cite{north-etal-2022-gmu} generates candidates using MLM and ranks them by MLM probability and word frequency. 
            \item \textbf{TSAR-LSBert} is a modified version of the original LSBert to support Spanish and Portuguese and produce more candidates. 
            \item \textbf{TSAR-TUNER} is an adaptive version of the TUNER system (a rule-based system) \cite{ferres_adaptable_2017} for the TSAR-2022 shared task.
        \end{itemize}
    
    \subsection{Evaluation Metrics}
        We adopted the same evaluation metrics used in TSAR-2022 shared task \cite{saggion-etal-2022-findings}. The metrics used are as follows:
        
        \begin{itemize}
            \item \textbf{Accuracy@1} (ACC@1): the percentage of instances with the top-ranked candidate in the gold candidates.  
            
            \item \textbf{Accuracy@N@Top1} (ACC@N@Top1): The percentage of instances where at least one of the top N predicted candidates match the most suggested gold candidates.
            
            \item \textbf{Potential@K}: the percentage of instances where at least one of the top K predicted candidates are present in the gold candidates.
            
            \item \textbf{Mean Average Precision@K} (MAP@K): the metric measures the relevance and ranking of the top K predicted candidates.  
        \end{itemize}

        To measure different aspects of the system's performance, we measured the results for different numbers of N and K candidates where N $\in$ \{1, 2, 3\} and K $\in$ \{3, 5, 10\}. ACC@1, MAP@1, and Potential@1 give the same results as per their definitions, so we report all of them as ACC@1 in the final results.
    
        
\subsection{Preprocessing} \label{sec: preprocessing}
            \begin{figure*}[ht]
                \begin{adjustbox}{width=1\textwidth}
                \centering
                \includegraphics{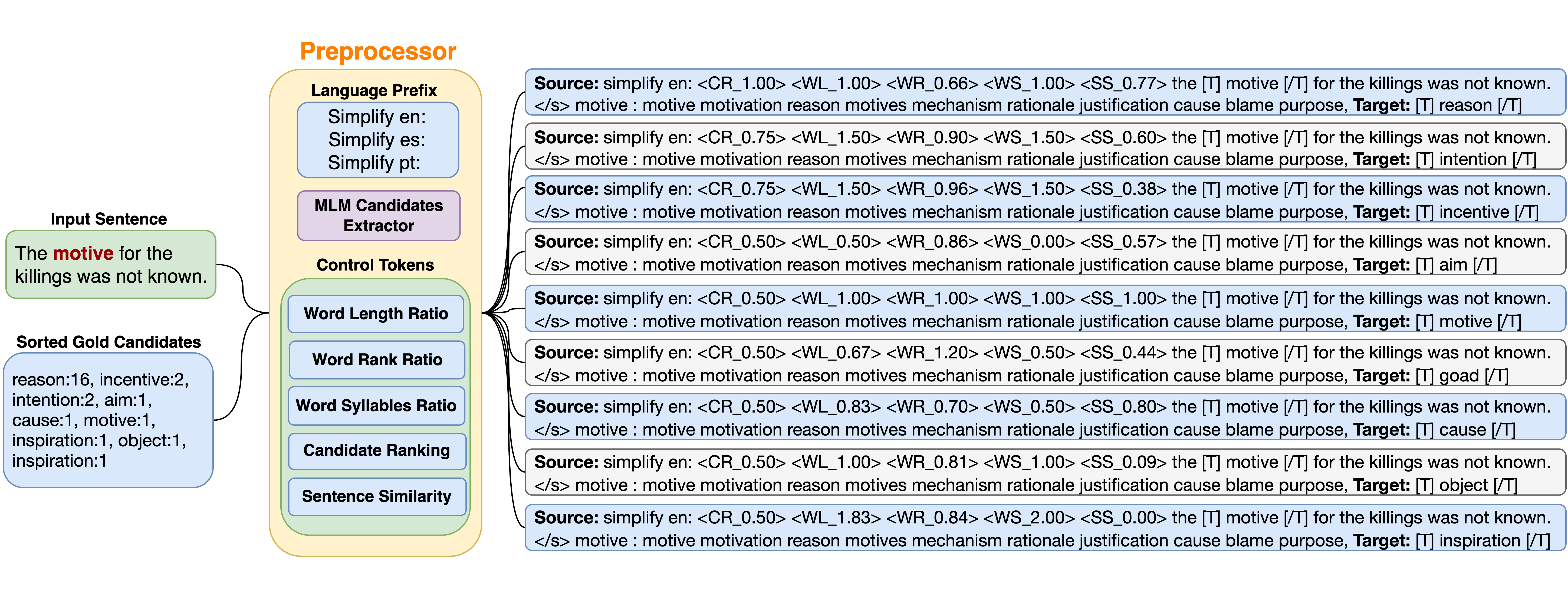}
                \end{adjustbox}
                \caption{Preprocessing steps of an English training example. For Spanish and Portuguese, the process follows the same procedures.}
                \label{figure:preprocessing_steps}
            \end{figure*}
            For each instance in the training set, there is a sentence, a complex word, and a list of ranked gold candidates. Thus, we compute the token values between the complex word and each candidate (we used all the candidates), which means if there are 9 candidates, there will be 9 training examples created.  

            Figure \ref{figure:preprocessing_steps} shows the preprocessing steps of an English sentence taken from the TSAR-EN dataset. The sentence contains the complex word "motive" and 9 ranked gold candidates; therefore, 9 training examples will be created. For each candidate and the complex word, we compute the tokens value, extract MLM candidates, and put all the values in the following format. Language prefix + Control Tokens + the input sentence with the complex word embedded in between [T] and [/T] + </s> + complex word + MLM candidates.   
            

            
            For Spanish and Portuguese datasets, we follow the same process and change the prefix to ``simplify es:'' for Spanish and ``simplify pt:'' for Portuguese.

            
                
                

            For the validation set, we follow the same format as the training set, except all the token values are set with the values of 1.00. E.g., <CR\_1.00> <WL\_1.00> <WR\_1.00> <WS\_1.00> <SS\_1.00>. We used these default values so that we could validate the model during the fine-tuning process and save the best model for evaluation.

            To choose the best pre-trained models for MLM candidates extraction, we ran a series of experiments on some of the most popular BERT-based pre-trained models (the popularity is based on the number of downloads available on Huggingface website\footnote{\url{https://huggingface.co/models}}). We compared them using the Potential metric since this metric measures the presence of the predicted candidates, which are matched with the gold candidates. For each model and each instance of a dataset, we extracted the top 10 candidates and computed the Potential. Table \ref{table:pretrained-model-tsar} in the Appendix reports the results of the TSAR dataset, and Table \ref{table:pretrained-model-others} in the Appendix shows the results of the LexMTurk, BenchLS, and NNSeval dataset. We did the experiments on the top 5, 10, 15, 20, 30, 40, and 50 candidates, and we found that the top 10 candidates worked the best in all of our experiments. So, these are the selected models that produce the best score in each dataset: ``roberta-base'' for TSAR-EN, ``PlanTL-GOB-ES/roberta-base-bne'' for TSAR-ES, ``neuralmind/bert-large-portuguese-cased'' for TSAR-PT, ``bert-large-cased'' for LexMTurk and BenchLS, and ``bert-base-uncased'' for NNSeval.

        \subsection{Model Details}
            
            In our experiments, we fine-tuned four different models: TLS-1, TLS-2, TLS-3, and mTLS. Each model was fine-tuned with the language prefix, control tokens, and MLM candidates, except for the TLS-3 model, which was without the MLM candidates.
           
            The following are the details of each model:
            \begin{itemize}
                \item TLS-1 is an English monolingual based on T5-large. It was fine-tuned and validated with the TSAR-EN dataset (we split the dataset to 80\% train, 20\% validation) and then tested with LexMTurk, BenchLS, and NNSeval. This model is intended to compare with LSBert and ConLS.
                
                \item TLS-2 is an English monolingual based on T5-large. It was fine-tuned, validated, and tested on the same dataset (TSAR-EN). The dataset was split into a 70\% train, a 15\% validation, and a 15\% test. 
                
                \item TLS-3 (without MLM candidates) is an English monolingual based on T5-large. It was fine-tuned, validated, and tested on the TSAR-EN dataset. The dataset was split into a 70\% train, a 15\% validation, and a 15\% test. 
                
                \item mTLS is a multilingual based on mT5-large. It was fine-tuned, validated, and tested with the whole TSAR-2022 dataset (TSAR-EN, TSAR-ES, TSAR-PT). We split the dataset of each language into a 70\% train, a 15\% validation, and a 15\% test. We then preprocessed, randomized, and combined the data of all languages into one training and one validation sets. During the fine-tuning process, the model is randomly fed with parallel data (the source and target data created by the preprocessing steps as shown in Figure \ref{figure:preprocessing_steps}) from the three languages, allowing the model to learn and share all the weights.

                \item The model TLS-2, TLS-3, and mTLS are intended to compare with the models from the TSAR-2022 shared task. In order to have a fair comparison between our model and the shared-task models, we only compared the results of the same 15\% test sets. 
                
            \end{itemize}

            We implemented our approach using Huggingface Transformers library\footnote{https://huggingface.co} and Pytorch-lightning\footnote{https://lightning.ai}. Then we fine-tuned each model on an NVidia RTX 3090 GPU with a batch size of 4 (except mTLS, the batch size was set to 1 due to out-of-memory issues), gradient accumulation steps of 4, max sequence length of 210 (it was based on the number of tokens/wordpiece from all datasets), learning rate of 1e-5, weight decay of 0.1, adam epsilon of 1e-8. We fine-tuned it for 30 epochs, and if the model did not improve for four epochs, we saved the best model based on the highest validation score ACC@1@Top1 and stopped the fine-tuning process. All of our models took less than 15 epochs to converge. We used a Python library called Optuna \cite{akiba_optuna_2019} to perform hyperparameters search on T5-small and T5-base to speed up the process and then employed the same hyperparameters in the final larger models like T5-large and mT5-large. For the generation, we used beam search and set it to 15 to generate 15 candidates so that it is left with around 10 candidates after some filtering (duplicate or the candidate the same as the complex word). In addition, in our experiments, the performance of the models based on T5-small and T5-base performed lower than the model based on T5-large in all metrics. The same with the multilingual models mT5-small, mT5-base, and mT5-large, so for that reason, we only report the results of the models that are based on T5-large and mT5-large.
            
        \subsection{At Inference}
            For each model, we performed a tokens value search on the validation set of each corresponding dataset using Optuna \cite{akiba_optuna_2019} (the same tool used for hyperparameters search). We searched the value of each token ranging between 0.5 and 2.0 with the step of 0.05, but we skipped the search for the Candidate Ranking token as we already knew the best value of it would be 1.00 to obtain the best candidates. We ran the search for 200 trials, then selected the top 10 sets of values that maximized ACC@1@Top1 and used them for the evaluation of the test set. For each set of tokens, we kept them fixed for all instances of the whole test set. Finally, we report the results of the set that maximized ACC@1@Top1. Figure \ref{figure:inference_example} shows an example from the TSAR-EN test set and the simpler substitutes generated by our TLS-2 model. 
            
            \begin{figure}[h]
                \centering
                \def\arraystretch{1.0}
                \begin{tabular}{p{0.94\linewidth}}
                \toprule
                \textbf{Source:} simplify en: <CR\_1.00> <WL\_1.25> <WR\_1.05> <WS\_1.60> <SS\_1.00> \#8-8 I want to continue playing at the highest level and win as many [T] \textbf{trophies} [/T] as possible. </s> \textbf{trophies} : trophies titles trophy competitions championships tournaments prizes awards cups medals \\
                \midrule
                \textbf{Predicted candidates:} awards,	medals,	prizes,	honors,	accolades,	titles,	crowns,	rewards,	achievements,	certificates \\
                \midrule
                \end{tabular}
                \caption{An example of the input taken from TSAR-EN test set and the candidates predicted by TLS-2 model.}
                \label{figure:inference_example}
            \end{figure}

\section{Results and Discussion} \label{sec: results and discussion}
    \begin{table*}[ht]
        \def\arraystretch{1.2}
        \begin{adjustbox}{width=1\textwidth,center=\textwidth}
        \begin{tabular}{llcccccccccc}
        \toprule
        \multirow{2}{*}{Dataset} & \multirow{2}{*}{System} & \multirow{2}{*}{ACC@1} & ACC@1 & ACC@2 & ACC@3 & MAP & MAP & MAP & Potential & Potential & Potential \\
         &  &  & @Top1 & @Top1 & @Top1 & @3 & @5 & @10 & @3 & @5 & @10 \\ \midrule
        \multirow{3}{*}{LexMTurk} 
         & LSBert & 0.8480 & 0.4400 & 0.5480 & 0.6040 & 0.5441 & 0.3901 & 0.2129 & 0.9320 & 0.9500 & 0.9580 \\
         & ConLS & 0.8060 & 0.4380 & 0.5639 & 0.6540 & 0.5545 & 0.4252 & 0.2759 & 0.9560 & 0.9820 & 0.9960 \\

         & \textbf{TLS-1} & \textbf{0.8580} & \textbf{0.4420} & \textbf{0.6040} & \textbf{0.7080} & \textbf{0.6567} & \textbf{0.5367} & \textbf{0.3572} & \textbf{0.9860} & \textbf{1.0000} & \textbf{1.0000} \\ \midrule

        \multirow{3}{*}{BenchLS} 
         & LSBert & 0.6759 & 0.4068 & 0.5145 & 0.5737 & 0.4229 & 0.2925 & 0.1574 & 0.8127 & 0.8428 & 0.8547 \\
         & ConLS & 0.6200 & 0.3799 & 0.5134 & 0.5931 & 0.4137 & 0.3054 & 0.1884 & 0.8127 & 0.8708 & 0.9031 \\
         & \textbf{TLS-1} & \textbf{0.7255} & \textbf{0.4133} & \textbf{0.5952} & \textbf{0.6749} & \textbf{0.5187} & \textbf{0.4015} & \textbf{0.2539} & \textbf{0.8848} & \textbf{0.9257} & \textbf{0.9612} \\ \midrule
        
        \multirow{3}{*}{NNSeval} 
         & LSBert & 0.4476 & 0.2803 & 0.3849 & 0.4393 & 0.2784 & 0.1997 & 0.1073 & 0.6485 & 0.7155 & 0.7448 \\ 
         & ConLS & 0.4100 & 0.2677 & 0.3430 & 0.4518 & 0.2731 & 0.203 & 0.1253 & 0.6109 & 0.6987 & 0.7908 \\
         & \textbf{TLS-1} & \textbf{0.5313} & \textbf{0.3263} & \textbf{0.4644} & \textbf{0.5397} & \textbf{0.3486} & \textbf{0.2762} & \textbf{0.1791} & \textbf{0.7824} & \textbf{0.8828} & \textbf{0.9414}
            \\
        \bottomrule
        \end{tabular}
        \end{adjustbox}
        \caption{Results of TLS-1 in comparison with LSBert and ConLS on the Accuracy@1, Accuracy@N@Top1, Potential@K, and MAP@K metrics. The best performances are in bold.}
        \label{table: results English monolingual}
    \end{table*}

    \begin{table*}[h]
        \def\arraystretch{1.2}
        \begin{adjustbox}{width=1\textwidth,center=\textwidth}
        \begin{tabular}{lccccccccccc}
        \toprule
        Model &
          \begin{tabular}[c]{@{}c@{}}ACC\\ @1\end{tabular} &
          \begin{tabular}[c]{@{}c@{}}ACC@1\\ @Top1\end{tabular} &
          \begin{tabular}[c]{@{}c@{}}ACC@2\\ @Top1\end{tabular} &
          \begin{tabular}[c]{@{}c@{}}ACC@3\\ @Top1\end{tabular} &
          \begin{tabular}[c]{@{}c@{}}MAP\\ @3\end{tabular} &
          \begin{tabular}[c]{@{}c@{}}MAP\\ @5\end{tabular} &
          \begin{tabular}[c]{@{}c@{}}MAP\\ @10\end{tabular} &
          \begin{tabular}[c]{@{}c@{}}Potential\\ @3\end{tabular} &
          \begin{tabular}[c]{@{}c@{}}Potential\\ @5\end{tabular} &
          \begin{tabular}[c]{@{}c@{}}Potential\\ @10\end{tabular} \\    
          \midrule
        TLS-2 & \textbf{0.8750} & \textbf{0.5536} & \textbf{0.6964} & 0.6964 & \textbf{0.6379} & \textbf{0.5126} & \textbf{0.3069} & 0.9643 & 0.9643 & \textbf{1.0000} \\
        TLS-3 & 0.8393 & \textbf{0.5536} & 0.6786 & \textbf{0.7500} & 0.5933 & 0.4506 & 0.2842 & 0.9643 & 0.9821 & 0.9821 \\
        mTLS & 0.6607 & 0.3929 & 0.5000 & 0.6071 & 0.4871 & 0.3651 & 0.2173 & 0.8571 & 0.9286 & 0.9643 \\ \midrule
        UniHD & \textbf{0.8750} & \textbf{0.5536} & 0.6429 & 0.6786 & 0.5913 & 0.4055 & 0.2284 & \textbf{1.0000} & \textbf{1.0000} & \textbf{1.0000} \\
        UoM\&MMU & 0.6964 & 0.4107 & 0.5536 & 0.5714 & 0.4315 & 0.3234 & 0.2020 & 0.8393 & 0.8571 & 0.8929 \\
        RCML & 0.6071 & 0.2321 & 0.4107 & 0.4821 & 0.3978 & 0.3032 & 0.1959 & 0.8214 & 0.9286 & 0.9464 \\
        LSBERT & 0.5893 & 0.2679 & 0.4821 & 0.5714 & 0.4385 & 0.3136 & 0.1860 & 0.8750 & 0.9107 & 0.9286 \\
        MANTIS & 0.5714 & 0.3036 & 0.4643 & 0.5179 & 0.4613 & 0.3463 & 0.2097 & 0.8393 & 0.9107 & 0.9464 \\
        GMU-WLV & 0.5179 & 0.2143 & 0.2500 & 0.4107 & 0.3700 & 0.2936 & 0.1716 & 0.7321 & 0.8393 & 0.9107 \\
        teamPN & 0.4821 & 0.1964 & 0.3571 & 0.3750 & 0.3065 & 0.2320 & 0.1160 & 0.6786 & 0.8036 & 0.8036 \\
        PresiUniv & 0.4643 & 0.1786 & 0.2857 & 0.3214 & 0.3075 & 0.2417 & 0.1396 & 0.6607 & 0.7500 & 0.7857 \\
        Cental & 0.4464 & 0.1250 & 0.2500 & 0.3393 & 0.3016 & 0.2210 & 0.1385 & 0.6607 & 0.7143 & 0.7857 \\
        CILS & 0.4107 & 0.1786 & 0.2500 & 0.2679 & 0.2817 & 0.2198 & 0.1378 & 0.5893 & 0.6071 & 0.6250 \\
        TUNER & 0.3929 & 0.1607 & 0.1607 & 0.1607 & 0.1865 & 0.1158 & 0.0579 & 0.4643 & 0.4643 & 0.4643 \\
        PolyU-CBS & 0.3571 & 0.1607 & 0.2321 & 0.3036 & 0.2579 & 0.1887 & 0.1118 & 0.6250 & 0.7500 & 0.8214 \\
        \bottomrule
        \end{tabular}
        \end{adjustbox}
        \caption{Official results from TSAR-2022 shared task in comparison with our models TSAR-EN dataset. The best performances are in bold.}
        \label{table:results_en}
    \end{table*}

    \begin{table*}[h!]
        \def\arraystretch{1.2}
        \begin{adjustbox}{width=1\textwidth,center=\textwidth}
        \begin{tabular}{lccccccccccc}
        \toprule
        Model &
          \begin{tabular}[c]{@{}c@{}}ACC\\ @1\end{tabular} &
          \begin{tabular}[c]{@{}c@{}}ACC@1\\ @Top1\end{tabular} &
          \begin{tabular}[c]{@{}c@{}}ACC@2\\ @Top1\end{tabular} &
          \begin{tabular}[c]{@{}c@{}}ACC@3\\ @Top1\end{tabular} &
          \begin{tabular}[c]{@{}c@{}}MAP\\ @3\end{tabular} &
          \begin{tabular}[c]{@{}c@{}}MAP\\ @5\end{tabular} &
          \begin{tabular}[c]{@{}c@{}}MAP\\ @10\end{tabular} &
          \begin{tabular}[c]{@{}c@{}}Potential\\ @3\end{tabular} &
          \begin{tabular}[c]{@{}c@{}}Potential\\ @5\end{tabular} &
          \begin{tabular}[c]{@{}c@{}}Potential\\ @10\end{tabular} \\    
          \midrule
              mTLS & \textbf{0.5357} & \textbf{0.2857} & \textbf{0.3929} & \textbf{0.4821} & \textbf{0.3790} & \textbf{0.2852} & \textbf{0.1685} & \textbf{0.7500} & \textbf{0.8036} & \textbf{0.9107} \\ \midrule
            PolyU-CBS & 0.4107 & 0.2143 & 0.2143 & 0.2143 & 0.2153 & 0.1479 & 0.0918 & 0.5000 & 0.5536 & 0.5893 \\
            GMU-WLV & 0.3929 & 0.1786 & 0.2679 & 0.3036 & 0.2560 & 0.1945 & 0.1167 & 0.5714 & 0.6607 & 0.7321 \\
            UoM\&MMU & 0.3571 & 0.1964 & 0.2679 & 0.3214 & 0.2391 & 0.1699 & 0.0979 & 0.5714 & 0.6250 & 0.7143 \\
            PresiUniv & 0.3214 & 0.1964 & 0.3214 & 0.3929 & 0.2361 & 0.1574 & 0.0860 & 0.6429 & 0.6786 & 0.7679 \\
            LSBERT & 0.3036 & 0.0893 & 0.1429 & 0.1786 & 0.1994 & 0.1504 & 0.0910 & 0.4643 & 0.6250 & 0.7500 \\
            Cental & 0.2679 & 0.1429 & 0.1786 & 0.2143 & 0.1865 & 0.1449 & 0.0851 & 0.5000 & 0.5536 & 0.5714 \\
            TUNER & 0.1429 & 0.0714 & 0.1071 & 0.1071 & 0.0843 & 0.0506 & 0.0253 & 0.1964 & 0.1964 & 0.1964
         \\
        
        \bottomrule
        \end{tabular}
        \end{adjustbox}
        \caption{Official results from TSAR-2022 shared task in comparison with our model on the TSAR-ES dataset. The best performances are in bold.}
        \label{table:results_es}
    \end{table*}
    
    
    \begin{table*}[h]
        \def\arraystretch{1.2}
        \begin{adjustbox}{width=1\textwidth,center=\textwidth}
        \begin{tabular}{lccccccccccc}
        \toprule
        Model &
          \begin{tabular}[c]{@{}c@{}}ACC\\ @1\end{tabular} &
          \begin{tabular}[c]{@{}c@{}}ACC@1\\ @Top1\end{tabular} &
          \begin{tabular}[c]{@{}c@{}}ACC@2\\ @Top1\end{tabular} &
          \begin{tabular}[c]{@{}c@{}}ACC@3\\ @Top1\end{tabular} &
          \begin{tabular}[c]{@{}c@{}}MAP\\ @3\end{tabular} &
          \begin{tabular}[c]{@{}c@{}}MAP\\ @5\end{tabular} &
          \begin{tabular}[c]{@{}c@{}}MAP\\ @10\end{tabular} &
          \begin{tabular}[c]{@{}c@{}}Potential\\ @3\end{tabular} &
          \begin{tabular}[c]{@{}c@{}}Potential\\ @5\end{tabular} &
          \begin{tabular}[c]{@{}c@{}}Potential\\ @10\end{tabular} \\    
          \midrule
        mTLS & \textbf{0.6607} & \textbf{0.4464} & \textbf{0.5536} & \textbf{0.5714} & \textbf{0.4216} & \textbf{0.2940} & \textbf{0.1842} & \textbf{0.8214} & \textbf{0.9107} & \textbf{0.9464} \\ \midrule
        GMU-WLV & 0.4464 & 0.2143 & 0.3750 & 0.4107 & 0.2579 & 0.1926 & 0.1143 & 0.6429 & 0.7679 & 0.8571 \\
        PolyU-CBS & 0.3571 & 0.1071 & 0.1429 & 0.1607 & 0.1905 & 0.1455 & 0.0847 & 0.4643 & 0.5536 & 0.6071 \\
        Cental & 0.3214 & 0.0714 & 0.1250 & 0.1964 & 0.2153 & 0.1554 & 0.0910 & 0.5714 & 0.6786 & 0.8214 \\
        LSBERT & 0.3036 & 0.1607 & 0.2321 & 0.3036 & 0.1895 & 0.1364 & 0.0816 & 0.5179 & 0.6250 & 0.7321 \\
        TUNER & 0.2321 & 0.1429 & 0.1607 & 0.1607 & 0.1071 & 0.0688 & 0.0344 & 0.2857 & 0.2857 & 0.2857 \\
        PresiUniv & 0.2321 & 0.1071 & 0.1786 & 0.1964 & 0.1409 & 0.0952 & 0.0532 & 0.3750 & 0.4643 & 0.5179 \\
        UoM\&MMU & 0.1071 & 0.0357 & 0.0536 & 0.0714 & 0.0704 & 0.0553 & 0.0338 & 0.1964 & 0.2500 & 0.2857 \\
        
        \bottomrule
        \end{tabular}
        \end{adjustbox}
        \caption{Official results from TSAR-2022 shared task in comparison with our model on TSAR-PT dataset. The best performances are in bold.}
        \label{table:results_pt}
    \end{table*}

    In our experiments, we compared our model with all the systems submitted to the TSAR-2022 shared task on the TSAR dataset and the other two state-of-the-art models, LSBert and ConLS, on LexMTurk, BenchLS, and NNSeval datasets. We compared all of them with the same metrics used in the TSAR-2022 shared task, such as ACC@1, ACC@N@Top1, Potential@1, and MAP@K where N $\in$ \{1, 2, 3\} and K $\in$ \{3, 5, 10\}.

    Table \ref{table: results English monolingual} presents the results of our model TLS-1 (a monolingual English model fine-tuned and validated on the TSAR-EN dataset) in comparison with LSBert and ConLS on LexMTurk, BenchLS, and NNSeval datasets.  Our model achieves better results in all metrics across the board, and the results on Potential@K and MAP@K show a significant improvement.

    Table \ref{table:results_en} shows the results of our three models, English monolingual models (TLS-2, TLS-3), and multilingual model (mTLS), compared with all the systems from the TSAR-2022 shared task on the TSAR-EN dataset. Since all the models from the shared task are unsupervised approaches, we only compare the results on the same 15\% test set. Our TLS-2 outperforms all the models in all metrics and performs equally to  GPT-3 model (UniHD) on ACC@1 and ACC@1@Top1, it also performs significantly better on ACC@\{2,3\}@Top1 and MAP@\{3,5,10\} but lower on Potential@\{3,5\}. 
    
    TLS-2 performs better than TLS-3 in all metrics except ACC@3@Top1, showing that adding MLM candidates does improve the model's performance.

    Our multilingual model (mTLS) performs better than the previous approaches, except for UniHD. The fact that the model's performance is notably inferior to its monolingual counterparts could be attributed to the following facts. First, the use of a multilingual model can reduce performance, as it contains a lot of irrelevant information from other languages. Second,  the mT5-large pre-trained model is significantly larger than the T5-large, with around 1.2 billion parameters compared to 737 million of the T5-large. Given the large number of parameters that need to be updated, the mT5-large model requires significantly more data to learn from; therefore, we could not fine-tune the mT5-large model individually for Spanish or Portuguese. We had to fine-tune a multilingual model (mTLS) by randomly feeding the data from the three languages, allowing the model to learn and share all the weights.

    Table \ref{table:results_es} and Table \ref{table:results_pt} present the results of our mTLS model in comparison with the TSAR-2022 official results on TSAR-ES and TSAR-PT datasets. Our model performs significantly better than all the participating systems in all metrics. However, there were unofficial results of UniHD that outperformed our mTLS model on TSAR-ES and TSAR-PT datasets.

\section{Conclusion and Future Work} \label{sec: conclusion}

    This paper proposed a new multilingual Controllable Transformer-based Lexical Simplification that integrates language-specific prefixes alongside dynamic control tokens and masked language model candidates to leverage the input-level information. This approach allows us to have the candidate generation and ranking within one model as well as multilingual. Moreover, our method enables the model to learn more effectively on the complex word and to have finer control over the generated candidates, leading the model to outperform all the previous state-of-the-art models in all datasets, including the GPT-3 model (UniHD) on some metrics. 

    For future work, we want to explore the use of large language models (LLMs) like LLaMA \cite{touvron_llama_2023} or MPT-7B\footnote{\url{https://www.mosaicml.com/blog/mpt-7b}} to perform instruction-based learning for Text Simplification. Recent work has shown that fine-tuning LLMs with instructions enables such models to achieve remarkable zero-shot capabilities on new tasks; this could have some potential for Text Simplification in situations where the training data is scarce. Moreover, since we only managed to assess the performance of our multilingual approach on a part of the TSAR-2022 corpus, we should explore ways to compare our trainable system with non-trainable ones in a more realistic setting.
    
\section*{Acknowledgements}
 We thank the anonymous reviewers for their constructive comments and suggestions. We acknowledge partial support from the individual project
Context-aware Multilingual Text Simplification (ConMuTeS)
PID2019-109066GB-I00/AEI/10.13039/501100011033 awarded
by Ministerio de Ciencia, Innovación y Universidades (MCIU)
and by Agencia Estatal de Investigación (AEI) of Spain. We also acknowledge support from  the project MCIN/AEI/10.13039/501100011033 under the Maria de Maeztu Units of Excellence Programme (CEX2021-001195-M) and  partial support from  Departament de Recerca i Universitats de la Generalitat de Catalunya.

\bibliographystyle{fullname}
\bibliography{Bibliography}

\appendix

            \begin{table*}[ht]
                \def\arraystretch{1.3}
                \begin{adjustbox}{width=1\textwidth,center=\textwidth}
                \begin{tabular}{lclclc}
                \toprule
                \multicolumn{2}{c}{TSAR-EN} & \multicolumn{2}{c}{TSAR-ES} & \multicolumn{2}{c}{TSAR-PT} \\
                Model & Potential & Model & Potential & Model & Potential \\ \midrule
                roberta-base & 0.971 & PlanTL-GOB-ES/roberta-base-bne & 0.837 & neuralmind/bert-large-portuguese-cased & 0.839 \\
                bert-large-uncased & 0.945 & PlanTL-GOB-ES/roberta-large-bne & 0.832 & neuralmind/bert-base-portuguese-cased & 0.811 \\
                bert-large-cased & 0.945 & dccuchile/bert-base-spanish-wwm-cased & 0.816 & xlm-roberta-large & 0.635 \\
                roberta-large & 0.943 & dccuchile/albert-xxlarge-spanish & 0.769 & xlm-roberta-base & 0.596 \\
                bert-base-uncased & 0.935 & dccuchile/albert-base-spanish & 0.738 & rdenadai/BR\_BERTo & 0.484 \\
                distilbert-base-uncased & 0.917 & dccuchile/distilbert-base-spanish-uncased & 0.664 & josu/roberta-pt-br & 0.461 \\
                bert-base-cased & 0.914 & xlm-roberta-large & 0.656 & bert-base-multilingual-cased & 0.386 \\
                albert-base-v2 & 0.867 & dccuchile/bert-base-spanish-wwm-uncased & 0.635 &  & \multicolumn{1}{l}{} \\
                xlm-roberta-large & 0.779 & bert-base-multilingual-uncased & 0.575 &  & \multicolumn{1}{l}{} \\
                 & \multicolumn{1}{l}{} & distilbert-base-multilingual-cased & 0.412 &  & \multicolumn{1}{l}{}\\ 
                 \bottomrule
                \end{tabular}
                \end{adjustbox}
                \caption{The comparison of different pre-trained models on candidate generation using masked language model ranked by Potential metric on TSAR dataset. Higher is better.}
                \label{table:pretrained-model-tsar}
            \end{table*}

            \begin{table*}[ht]
                \def\arraystretch{1.1}
                \begin{adjustbox}{width=1\textwidth,center=\textwidth}
                \begin{tabular}{lclclc}
                \toprule
                \multicolumn{2}{c}{LexMTurk} & \multicolumn{2}{c}{BenchLS} & \multicolumn{2}{c}{NNSeval} \\
                Model & Potential & Model & Potential & Model & Potential \\ 
                \midrule
                bert-large-cased & 0.974 & bert-large-cased & 0.918 & bert-base-uncased & 0.887 \\
                bert-base-uncased & 0.972 & bert-large-uncased & 0.909 & roberta-base & 0.883 \\
                bert-large-uncased & 0.970 & roberta-base & 0.906 & bert-large-uncased & 0.879 \\
                roberta-base & 0.970 & bert-base-uncased & 0.899 & bert-base-cased & 0.870 \\
                bert-base-cased & 0.962 & bert-base-cased & 0.893 & bert-large-cased & 0.858 \\
                distilbert-base-uncased & 0.950 & distilbert-base-uncased & 0.869 & distilbert-base-uncased & 0.791 \\
                xlm-roberta-large & 0.934 & albert-base-v2 & 0.850 & albert-base-v2 & 0.762 \\
                albert-base-v2 & 0.926 & roberta-large & 0.830 & roberta-large & 0.745 \\
                roberta-large & 0.904 & xlm-roberta-large & 0.813 & xlm-roberta-large & 0.711 \\ 
                \bottomrule
                \end{tabular}
                \end{adjustbox}
                \caption{The comparison of different pre-trained models on candidate generation using masked language model ranked by Potential metric on LexMTurk, BenchLS, and NNSeval dataset. Higher is better.}
                \label{table:pretrained-model-others}
            \end{table*}

\end{document}